**Stimulus Motion Perception Studies Imply Specific Neural Computations in Human Visual Stabilization**

*David W Arathorn[1], Josephine C. D'Angelo[2], Austin Roorda[3]*

*(Version: 6/22/2026)*

**Abstract:**

Even during fixation the human eye is constantly in low amplitude motion, jittering over small angles in random directions at up to 100Hz. This motion results in all features of the image on the retina constantly traversing a number of cones, yet objects which are stable in the world are perceived to be stable, and any object which is moving in the world is perceived to be moving. A series of experiments carried out over a dozen years revealed the psychophysics of visual stabilization to be more nuanced than might be assumed, say, from the mechanics of stabilization of camera images, or what might be assumed to be the simplest solution from an evolutionary perspective. The psychophysics revealed by the experiments strongly implies a specific set of operations on retinal signals resulting in the observed stabilization behavior. The presentation is in two levels. First is a functional description of the action of the mechanism that is very likely responsible for the experimentally observed behavior. Second is a more speculative proposal of circuit-level neural elements that might implement the functional behavior.

**Introduction:**

This paper presents a possible neural mechanism to explain the results of a sequence of psychophysical experiments conducted over the course of a dozen years part of whose objective has been to understand the mechanism which allows us to see stable images despite the significant jitter of eye motion during fixation. The first of these experiments, reported in 2013 [1] exploited the newly developed capability to deliver laser stimuli to the retina with less than cone-diameter accuracy entirely contingent on the motion of the eye itself [2]. This new capability revealed an unexpected phenomenon: a moving stimulus appears stable when its motion on the retina is moving opposite to eye motion, largely independent of the amplitude of that motion, but appears to move as a projection of world motion when moving in the same direction as eye motion, and of course appears stable when not moving in the world frame. This effect varies with the angle of stimulus motion with respect to eye motion [1]. This result, of course, implied that whatever neural mechanism is responsible for stabilizing the stimulus image depends on that mechanism having knowledge of the direction and amplitude of eye motion. This raised the question of where the responsible neural mechanism obtains that information: from the retinal image or from a motor signal corresponding to the net result of oculomotor commands, the so-called "efferent copy." Because of limitations in the technology used in the first experiments, the answer to this remained uncertain. However, with dramatic improvements to both the technology and the visual presentation technique, a set of experiments performed in 2024 and 2025 [3, 4] has provided

[1] Montana State University, Dept of Electrical and Computer Engineering, dwa@cns.montana.edu (corresponding)
[2] University of California Berkeley, Herbert Wertheim School of Optometry and Vision Science
[3] University of California Berkeley, Herbert Wertheim School of Optometry and Vision Science and School of Optometry and Vision Science, University of Waterloo, Ontario, Canada

a robust answer to that question. Once again, the answer turns out to be more nuanced than expected. The results of the combined experiments have narrowed the possibilities enough to allow a functional circuit level hypothesis as to the neural mechanism responsible for the observed psychophysics. That is the subject of this paper.

The experimental results reported in papers subsequent to [1] only tested stimulus motion along the axis of eye motion, and the neural mechanism proposed here appears to be a parsimonious explanation of the source of that behavior. The results in [1] of tests of stimulus motion over the full range of angles away from the axis of eye motion, while not as extensive as results in the later papers, were strongly suggestive of one or more plausible modes of operation of the same neural mechanism. It is likely that the characteristics of the proposed mechanism imply behaviors which have not yet been practical to test experimentally, but may be inferred.

The term “functional” implies that the experimentally observed behavior is explained in terms of the sequence of operations of chunks of neural circuitry which perform actions like *matching* (performing the rough equivalent to a dot product of two arrays or vectors of signals) and *mapping* (performing the strict equivalent of a geometrical shift or translation of one neural array to another. A short discussion of how such chunks of neural circuitry may work at a signal level follows the main functional discussion but should be taken as one possible implementation. The presentation is primarily textual and diagrammatic.

**The Problem:**

What is the nature of the neural circuitry implied by the fact that we can see a stable image despite the unquestionable fact that the image on the retina is anything but stable due to continual motion of the eye even when fixating? Due to the high frequency of the eye’s motion this is a more extreme version of problem encountered by small video cameras held by shaky hands, but on the identical scale as the problem encountered by the AOSLO (adaptive optics scanning laser ophthalmascope) instrument used to obtain precise stable images of the retinal mosaic despite the motion of the eye. The operation common to cameras and AOSLO is *mapping*: i.e. the deliberate geometric shift or translation (and if necessary rotation) of the retinal image at any moment to align it with a previously captured image of the same scene. The parameters of the mapping (the direction and magnitude of the translation (and angle of rotation if necessary) may be obtained in different ways, but the operation that keeps the image stable over time is always simply mapping.

The use of the term “retina” (or *retinal spatial array*) in this discussion is not literal but rather refers to the last stage of the pathway from the physical retina which preserves the spatial ordering of the physical retina and is considered here to be the stage in the processing stream to which mapping is applied. The image that is perceived by the observer is considered to be held in a spatial array termed here, for lack of neuro-anatomical suspects, the *canvas*. Two other neural spatial arrays play a role in the process: the *background reference spatial array* and the *stimulus reference spatial array*.

The problem posed by experiments to determine the character of human visual stabilization is that, unlike the optical instruments, it is not simply a matter of constant application of the correct mapping to compensate eye motion and keep the image still for the observing brain. This paper presents a necessarily more complex mode of application of the basic mapping operation necessary to explain the observed psychophysics. The latter is the *functional* description, and the primary goal of this paper. Secondarily, the paper also presents a basic neural circuit sufficient to accomplish the necessary mapping operation.

For the moment we simply assume there exists a neural circuit that

1) performs an operation which concurrently compares the “retinal” spatial array at a large number of different translations with one or more reference spatial arrays and activates a neuron which corresponds to the translation which produce the best match between the retinal array and the particular reference array. (For readers with some background in image processing, this corresponds to image cross-correlation by 2D FFT-IFFT, wherein the element of the output array with the highest value indicates the magnitude and direction of the best translation, and the value of that element indicates the degree of match.)
2) causes the neuron which corresponds to the current best match to activate the actual transfer of the correspondingly translated “retina” spatial array into a canvas spatial array, replacing the data that was previously located there. The activation pattern of the canvas spatial array is what is perceived. This implies that there must be several pathways: one for establishing the match with each reference spatial array, and the other for updating the canvas spatial array.

Of course it isn’t plausible that the brain computes FFTs and iFFTs. However there is a simple neuronal circuit with the above functional characteristics, which will be described in more detail later in this paper. The pathways between the spatial arrays that implement the required behavior are shown in Figure 1. The various modes of function of this mechanism are analyzed later in this paper.

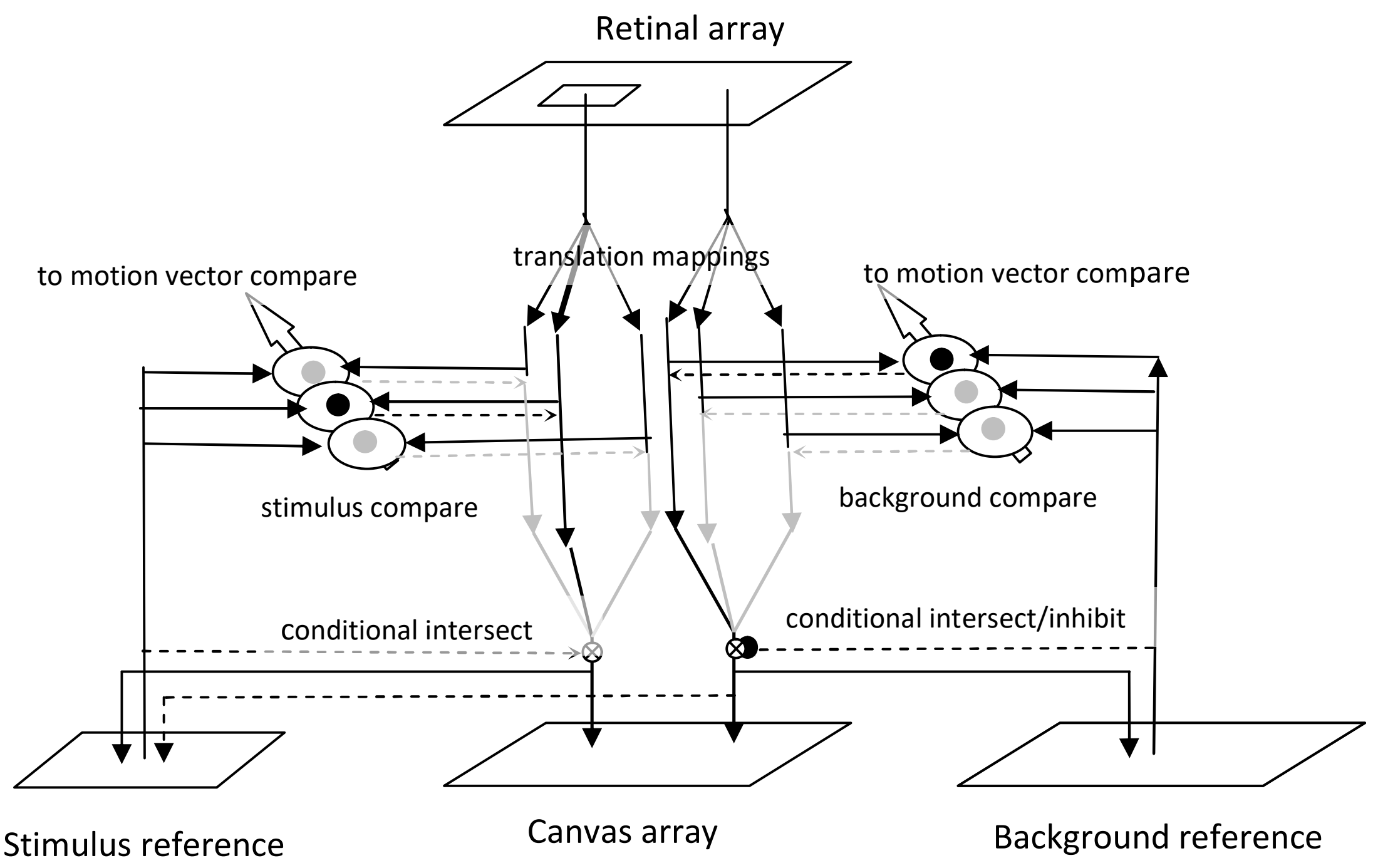

**Figure 1:** Proposed visual stabilization pathways

**Experimental conditions:**

In the experiments for antecedent papers [1,3,4] a stimulus pattern is moved by the instrument contingent on the eye motion as measured by the instrument at a frequency of about 1000Hz. The position of the stimulus in the display is updated at 60Hz. The direction and amplitude of the stimulus motion is controlled by a gain factor which scales the measured eye motion and by its sign determines the direction of stimulus motion relative to the motion of the physical retina. (In the first experiments, an angle parameter also altered the direction of stimulus motion relative to eye motion, whereas in later experiments only motion in the axis of eye motion is considered.) The definition of gain is such that gain = 0.0 means the stimulus does not move under control of eye motion and is hence fixed in the "world." Gain = 1.0 means the stimulus moves along with eye motion and is hence in a fixed location on the retina as the eye moves. Gains > 1.0 mean stimulus moves ahead of eye motion on the retina. Gains < 0.0 mean stimulus moves contrary to eye motion, hence very quickly on the retina. Gains between 0.0 and 1.0 mean stimulus is moving in the direction of eye motion but slower than the retina, so falling behind the motion of the retina. In other words, for the range of gains < 1.0, the image moves in the direction of the normal retinal slip. Over the range of gains differing percepts of stimulus stability are elicited and hence reveal stabilization behavior more complex that the stabilization in cameras or the AOSLO instrument. Also the presence or absence of background pattern for reference dramatically affects stabilization behavior.

**Summary of results from the earlier papers:**

In the experiments *background* is a pattern which extends over a wide area of the retina and of course is stable in the world. *Stimulus* is a very small filled circle located either in or near the fovea which is moved under instrument control contingent on the motion of the eye, and is therefore moving in the world frame unless the gain is set to zero.

In presentations with background the background is always perceived as stable, while the stimulus is perceived as stable or nearly stable for gains less than 1.0, but is perceived as moving with approximately world motion for gains greater than 1.0.

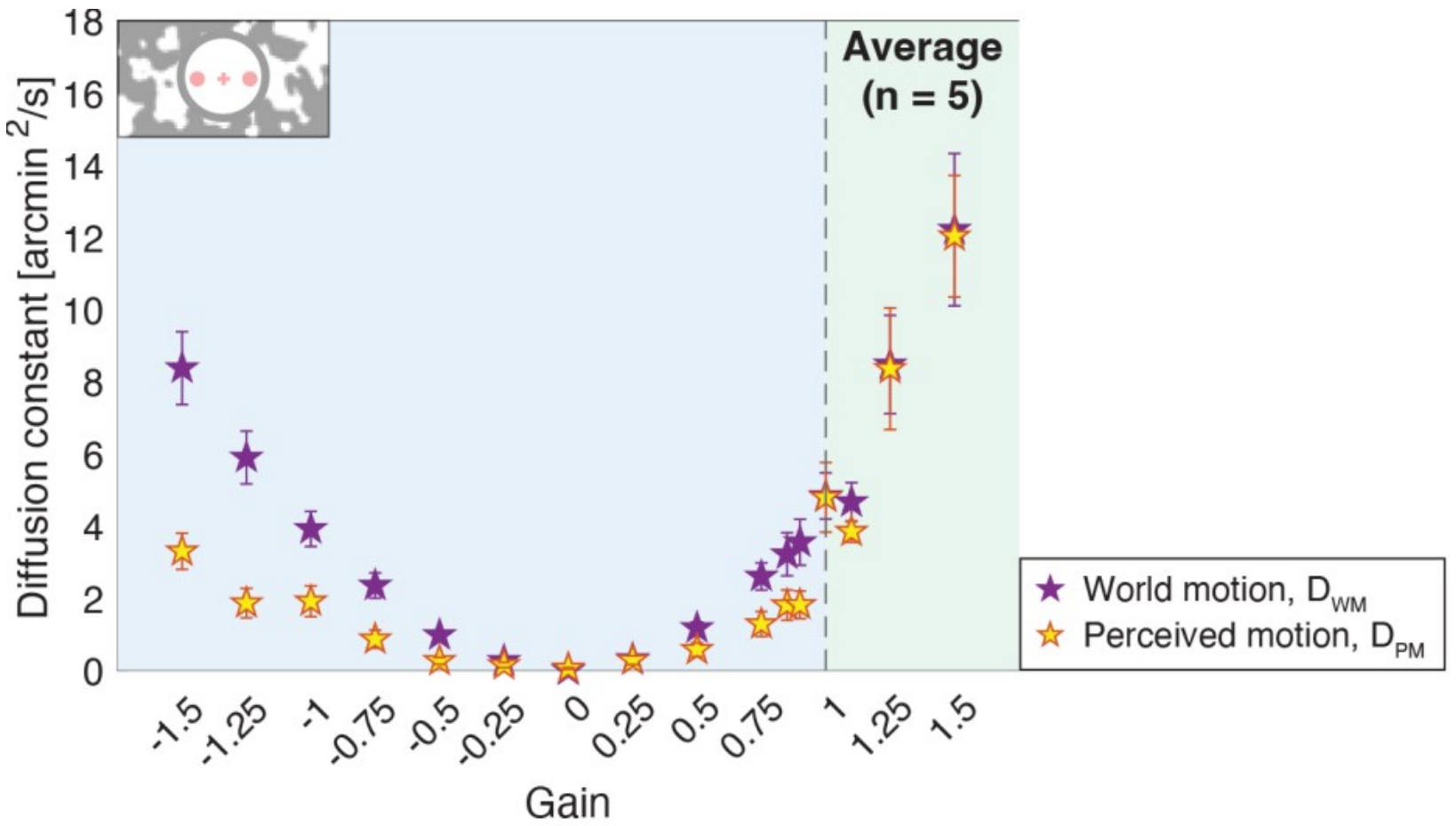

**Figure 2:** Relationship of stimulus motion in world frame and perceived motion as a function of gain when background is present (averaged over several subjects). From [4]: Appdx 1, A1.2.

The graph above is data averaged from several subjects. The individual differences are important and will be discussed later. (The reader is referred to [3, 4 ] for an explanation of how the above data is acquired and processed.)

In presentations with no background the stimulus is perceived to move with some magnitude which is approximately proportional to either the world motion or the motion on the retina.

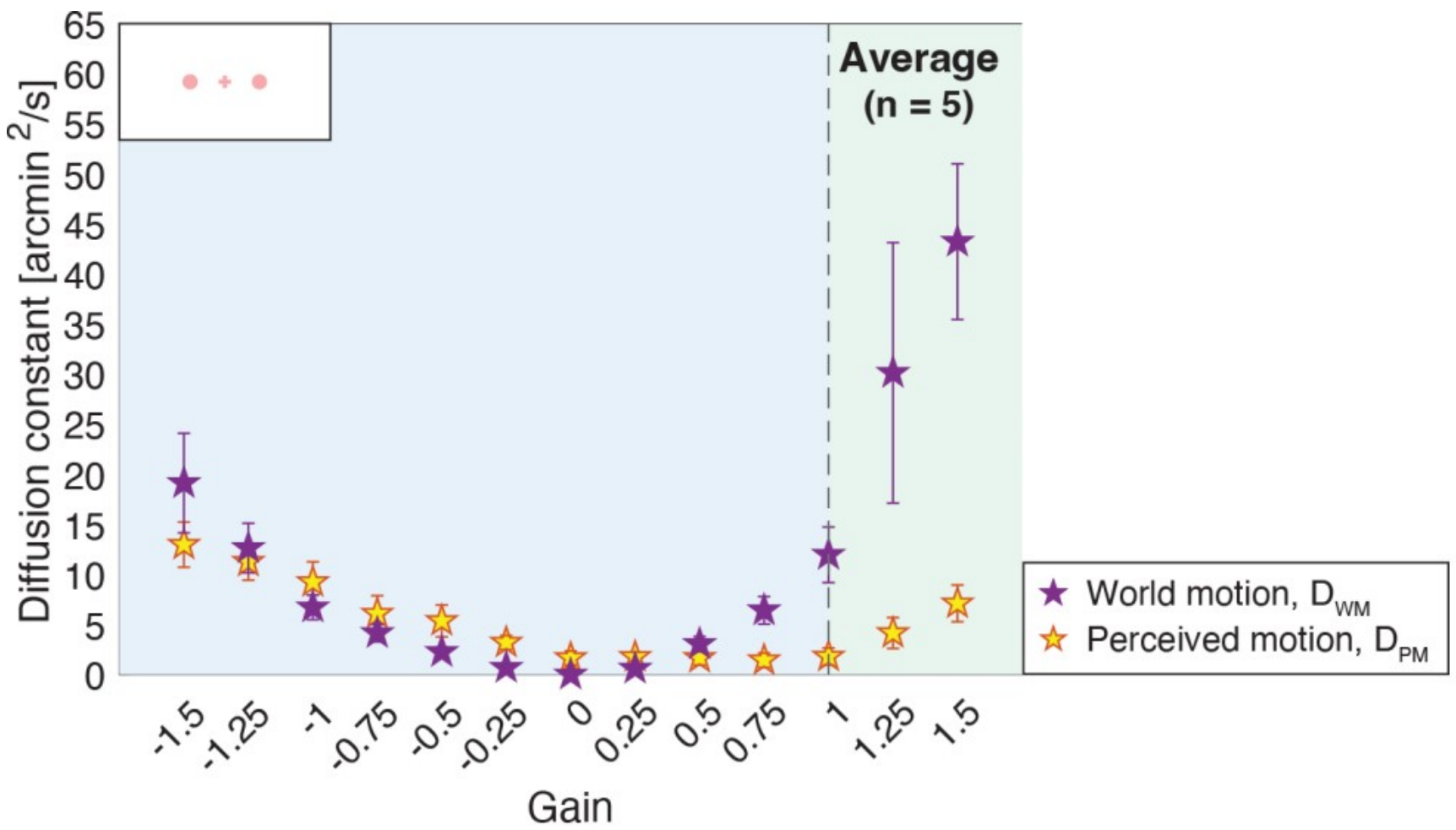

**Figure 3:** Relationship of stimulus motion in world frame and perceived motion as a function of gain when background is absent (averaged over several subjects). From [4]: Appdx 1, A1.7.

The graph above is data averaged from several subjects. The individual differences are important and will be discussed later. (The reader is referred to [3, 4] for an explanation of how the above data is acquired and processed.)

The behavior of the neural mechanism that accounts for the above results is summarized as follows. We assume that the background reference array and stimulus reference array are initialized at the beginning of each fixation and remain unchanged until a saccade or change of attention. The background (if present) and stimulus are separated by inhibitive subtraction as the stimulus moves relative to the background. Then as the eye moves the retinal array is compared with the reference array to determine the direction and amount of shift of both the background (if present) and the stimulus.

**Percepts when Background is Present:**

Background, if present, establishes primary mapping. The measured shift between retinal array and background reference array causes the retinal array to be shifted by the same degree and transferred to the canvas array, so that the background is perceived not to move, though any small changes are visible. As the eye moves the measured shift changes and the background pattern continues to be transferred to the canvas array so that it appears to be stable.

**If the stimulus has not moved relative to the background** (i.e. gain = 0.0) the same mapping registers background and stimulus. This is consistent with the report of zero perceived motion (or very near) for gain = 0.0 stimuli by all subjects.

**If stimulus has moved relative to background**, then a secondary mapping registers (measures shift) of stimulus region of retinal image and the position of the stimulus pattern in the stimulus reference array. Because the primary mapping executes first, there is a small latency in the transfer of retinal image of the stimulus via the secondary mapping, if it occurs. This delay is necessary to allow the background surrounding the stimulus to be masked. This latency becomes significant at high stimulus velocities.

**If direction of stimulus mapping is consistent with the direction of the background mapping** (same direction relative to retina, or within stabilization sector as in earlier paper [1]) **then stimulus region of retinal image is actually transferred onto canvas array via the stimulus mapping.** The stimulus therefore appears in the canvas array at the same relative location it resides in the stimulus reference array, and therefore appears to be stabilized (with minor deviances under some conditions.)

For small gains, and/or low eye motion velocities, there is negligible displacement of stimulus within time window, so mapping results in accurate location of transfer of stimulus to canvas stimulus region. This results in stimulus appearing to be stable despite veridical motion relative to background. For large displacements due to high negative gain or occasional high eye motion velocity, the latency of the transfer means mapping does not fully compensate stimulus motion, so residual shift (world motion minus mapping shift) of stimulus on canvas results and is perceived as minor stimulus motion.

An important feature of the background-present data is the sharp discontinuity in perceived motion at or near gain = 1.0, as reported in [4]. This is the gain threshold below which the stimulus is moving less than or opposite to eye motion and above which it is moving ahead of eye motion. This discontinuity indicates a change in the mode of mapping.

**If direction of stimulus mapping is inconsistent** (different direction or outside stabilization sector) **with the direction of background mapping then the background mapping overrides the stimulus mapping and transfers the current retinal image including stimulus to reference background with perceptual result that stimulus moves with world displacement relative to background.** Any changes to background or stimulus patterns are transferred to canvas and will be perceived.

The above behavior is evident in the data for individual subjects observing with background-present.

For a subject with very small fixational eye movements the stabilization for gains < 1.0 is very effective.

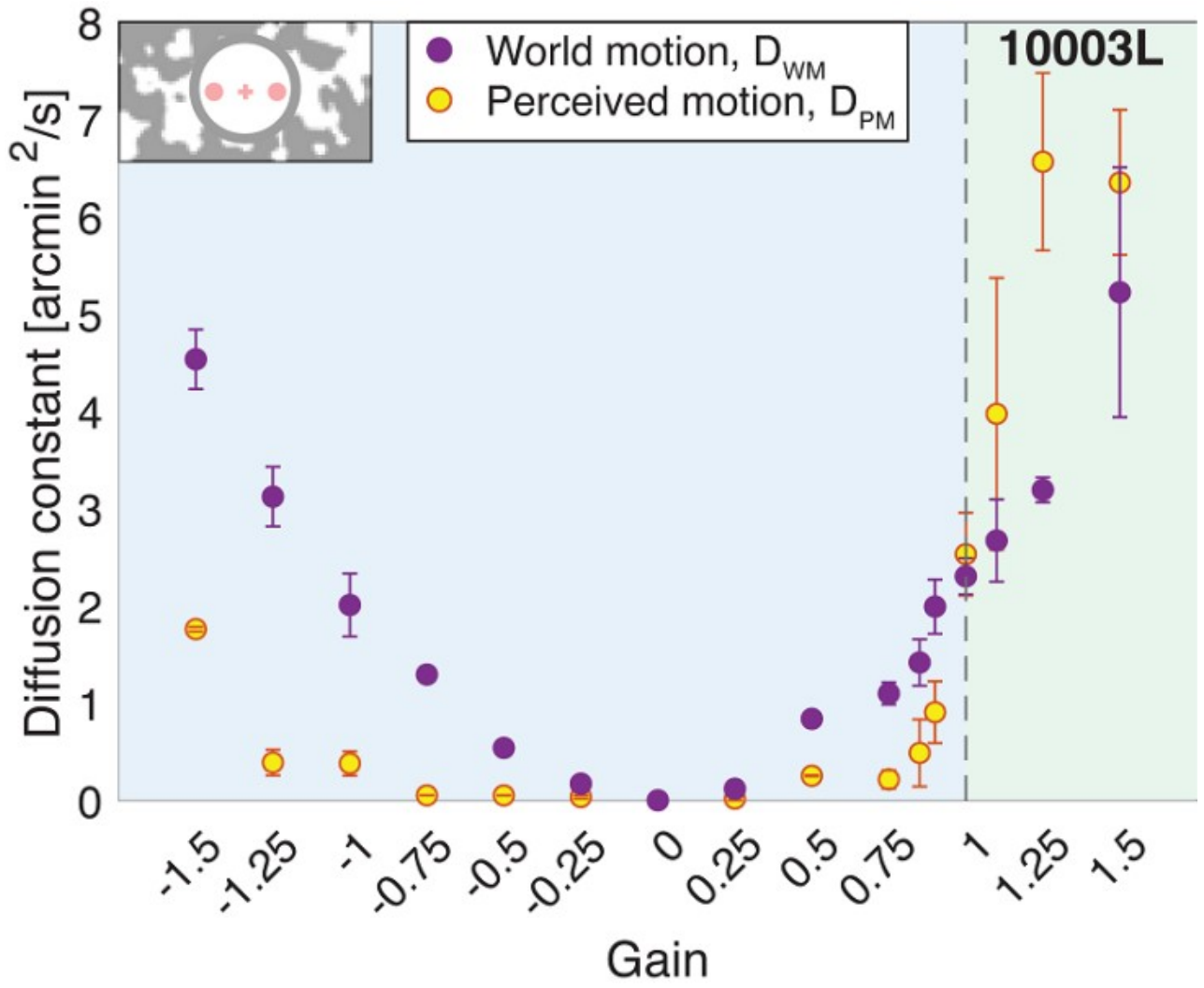

**Figure 4:** Relationship of stimulus motion in world frame and perceived motion as a function of gain when background is present for subject with strong fixation. From [4]: Figure 2C.

For a subject with larger fixational eye movements the stabilization for gains < 1.0 is very effective for small gains but becomes less effective as the gain, hence stimulus motion, increases

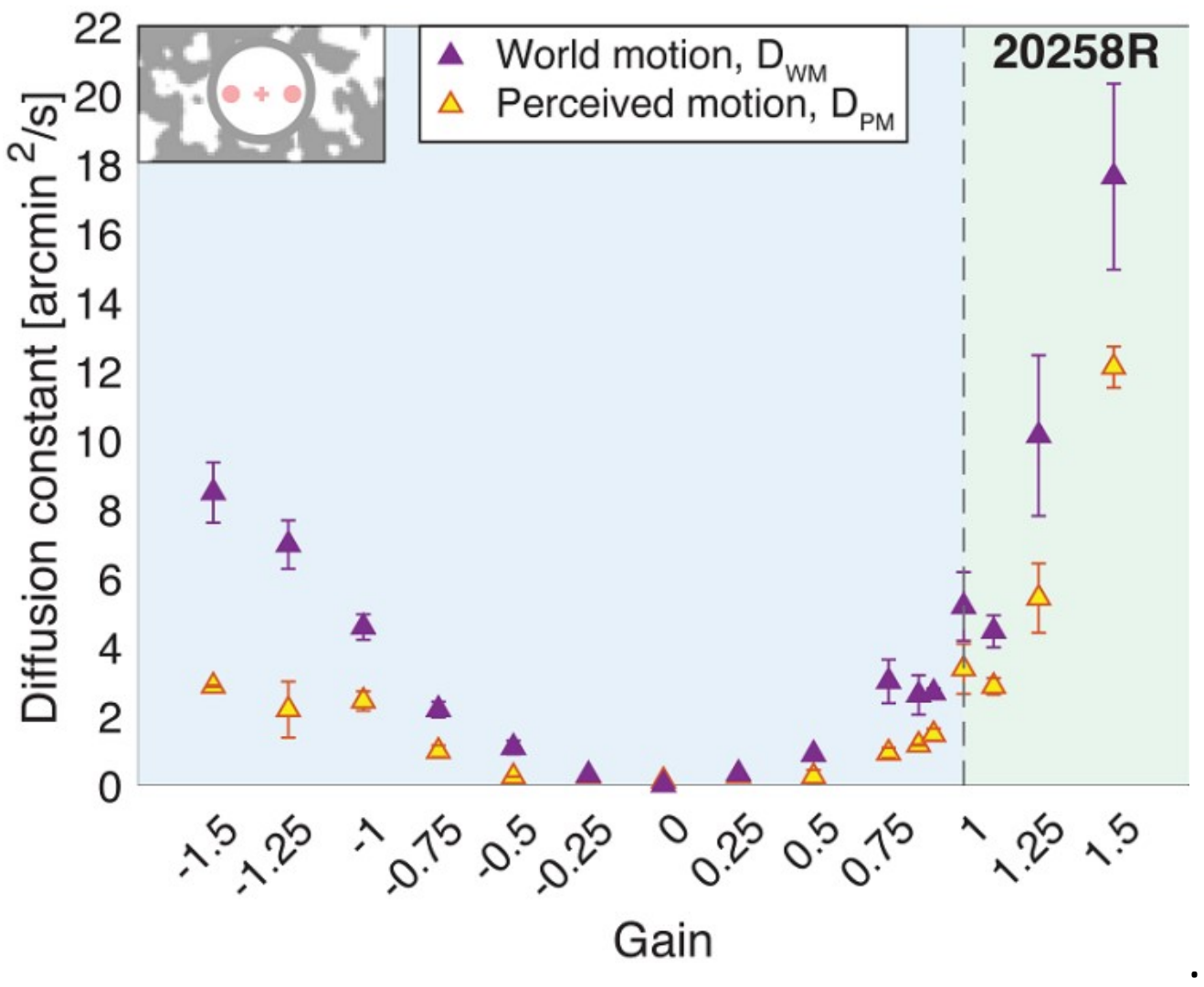

**Figure 5:** Relationship of stimulus motion in world frame and perceived motion as a function of gain when background is present for subject with weaker fixation. From [4]: Appdx 1, A1.2.

The decrease in perceived stability for negative gains further from 0.0 seem to be best explained by the latency between the primary and secondary mapping operations to the canvas array. If we assume a constant latency, then the faster the stimulus moves on the retina the more it will have moved on the canvas array before the next secondary mapping restores it to its reference position. This is perceived on the canvas array as an increasing degree of instability for larger negative gains.

The divergence from zero perceived motion for positive gains just below 1.0 can't be explained by latency because the motion of the stimulus on the retina is very small. More likely this is a consequence of occasional asymmetric jitter in the transition threshold between mapping modes near gain = 1.0: i.e. sometimes in motion-suppressed mode and sometimes in veridical motion mode.

The gain range around gain = 0.0 for which stabilization is perfect or near perfect is what one would expect for a system which, evolving in the natural world, would provide the most useful percepts in normal operating conditions. That this range applies to stimuli as well as background suggests that the stimulus mapping mechanism inherited its range from the background mapping but had to be "pruned" to provide a regime in which stimuli moving in world relative to background were correctly perceived.

**Percepts when Background is Absent (Ganzfeld):**

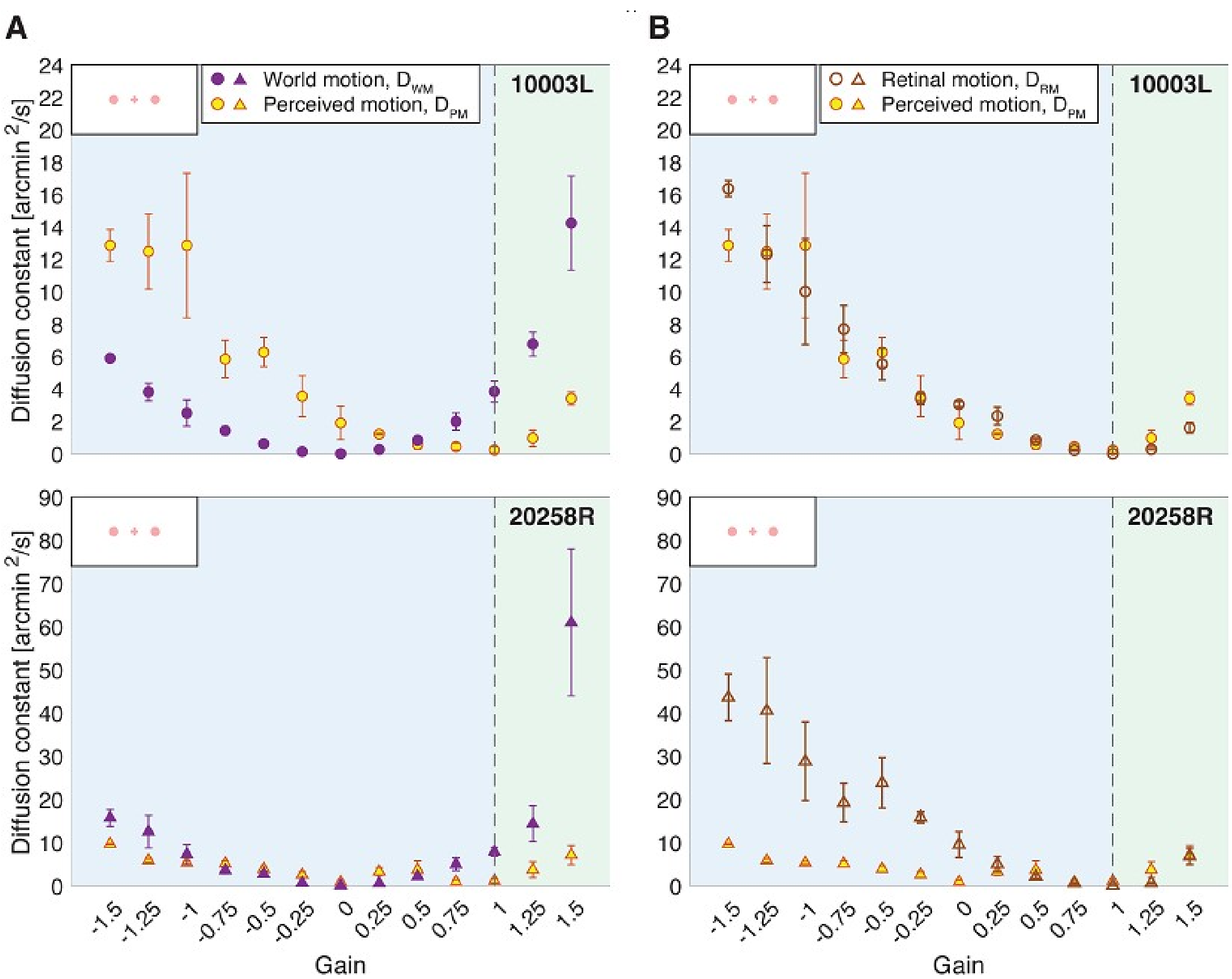

**Figure 6: A**: Relationship of stimulus motion in *world* frame to perceived motion. **B**: relationship of *retinal* motion to perceived motion. Background is absent for all cases. Subjects with strong fixation (top) and weaker fixation (bottom). From [4]: Appdx 1, A1.7.

.

For small eye motion subjects (e.g. 10003) there is no background motion information available so the primary mapping defaults to the identity mapping so that the current retinal image (or subsets) are transferred to the canvas, resulting in perception of retinal motion of any stimuli.

For large eye motion or weaker fixation subjects (Figure 6A bottom),, the data indicate the perceived motion of the stimulus is approximately the world motion, meaning that eye motion has been subtracted. **This implies that for large eye motions there is likely an efferent signal approximating eye motion** which can drive the mapping selection to compensate the eye motion. Given the complex relationship of net eye rotation and the individual oculomotor control signals, the generation of an efferent signal representing net eye rotation is likely to be approximate. By comparison the evidence from a subject with strong fixation (Figure 6B top), hence limited eye motion amplitude, strongly indicates a fixed or identity mapping, any efferent signal corresponding to eye motion either does not exist or is ignored by the mapping mechanism.

The background-absent condition is an anomaly in the real world. It is equivalent to observing a bird in flight directly overhead against a clear sky from a mountain top, such that no fixed background falls on the retina. Since in this condition the bird image does not jitter some signal must drive the compensation for eye motion. One might be inclined to conclude this could be the result of a low resolution efferent copy signal. However, the fact that an isolated bright light in a darkened room often appears to wander, suggests that no effective stabilizing signal is available. Other explanations of this dichotomy may be relevant.

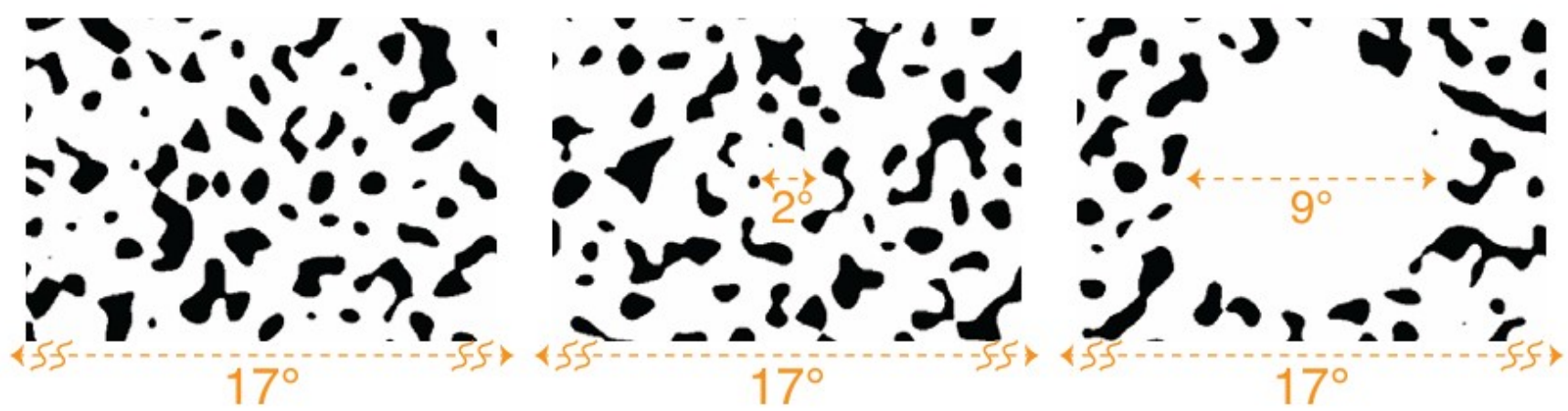

**Figure 7:** Different background extent conditions tested to determine effect on stabilization behavior

Under controlled experimental conditions, the presence of a background field only far in the periphery (as produced by a 9.0 degree white circle, Figure 7) appears to produce a perceptual behavior approaching that produced by the complete absence of background [4]. This result suggests that background beyond a certain radius on the retina either does not contribute at all (or significantly due to lower cone density) to the matching operation, and hence does not invoke background mapping (or does so inconsistently) and consequently the primary mapping defaults to the identity map, giving rise to percepts similar to complete absence of background.

**Numerical results:**

Comparison of experimental and simulated world motion versus perceived motion shows a close match, Figure 8. The software that implements this follows the computation described later for Figures 1b-e and is driven by recorded eye motion data for the same subject. Simulation does not include any latency artifacts so the residual perceived motion at high negative gains is absent from the model results. To represent those accurately would require knowledge of the dynamics of the as yet unidentified neural circuitry responsible for the stabilization.

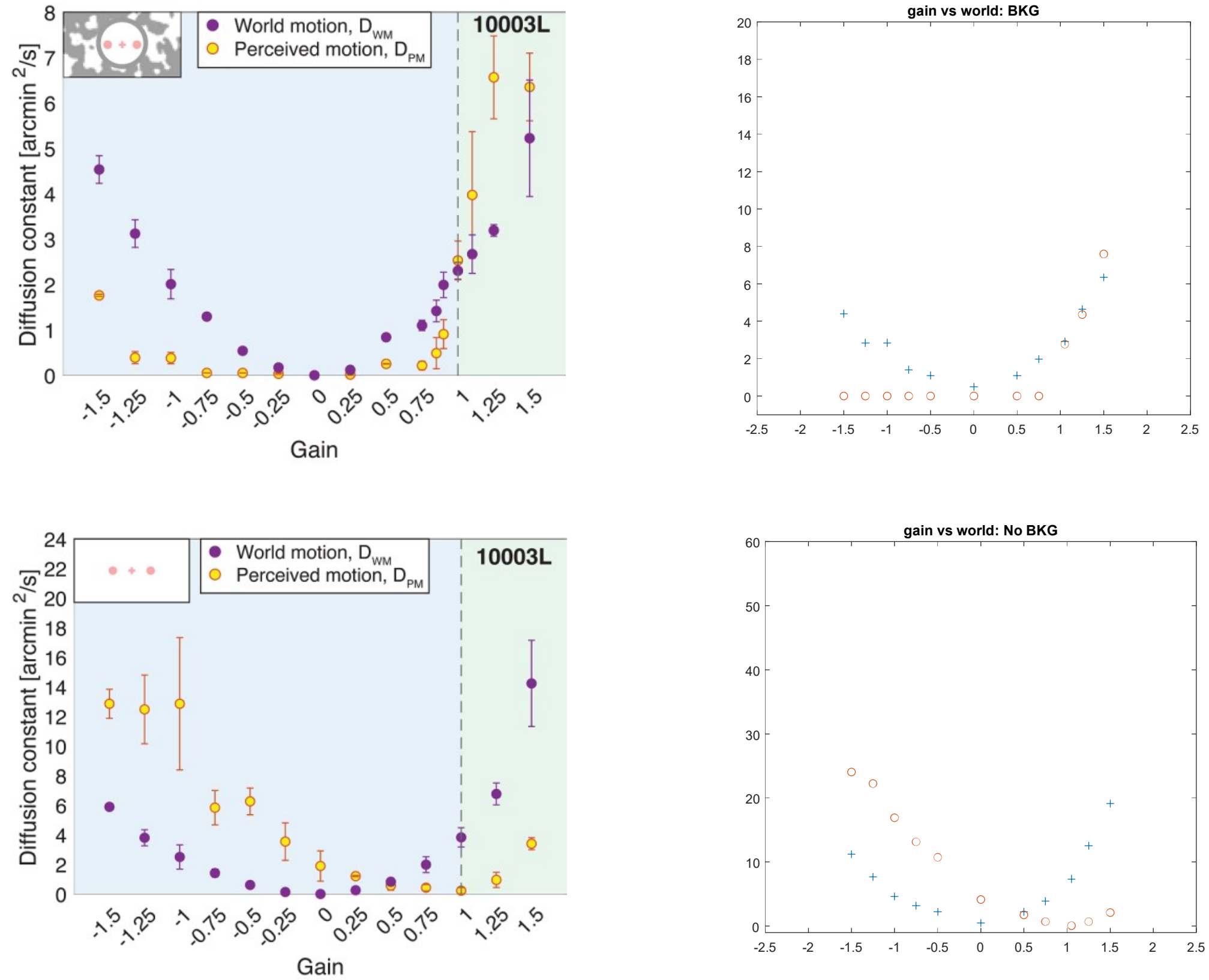

**Figure 8:** Top pair: Relationship of stimulus motion in world frame and perceived motion as a function of gain when *background is present* for subject with strong fixation versus numerical simulation. Bottom pair: Relationship of stimulus motion in world frame and perceived motion as a function of gain when *background is absent* for subject with strong fixation versus numerical simulation. Experimental data (left) from [4]: Figure 2C and [4]: Appdx 1, A1.2. In both simulations (top, bottom right): '+' world motion, 'o' percept.

**Psychophysical Behavior, Background-Present:**

A graphical presentation clarifies the relationship between world motion of the background and stimulus, and the resulting retinal motion and subject percept. The hexagon represents background pattern, always fixed world position. The eye motion is rightward and downward. Circle with cross is stimulus whose motion is contingent on eye motion with different gains.

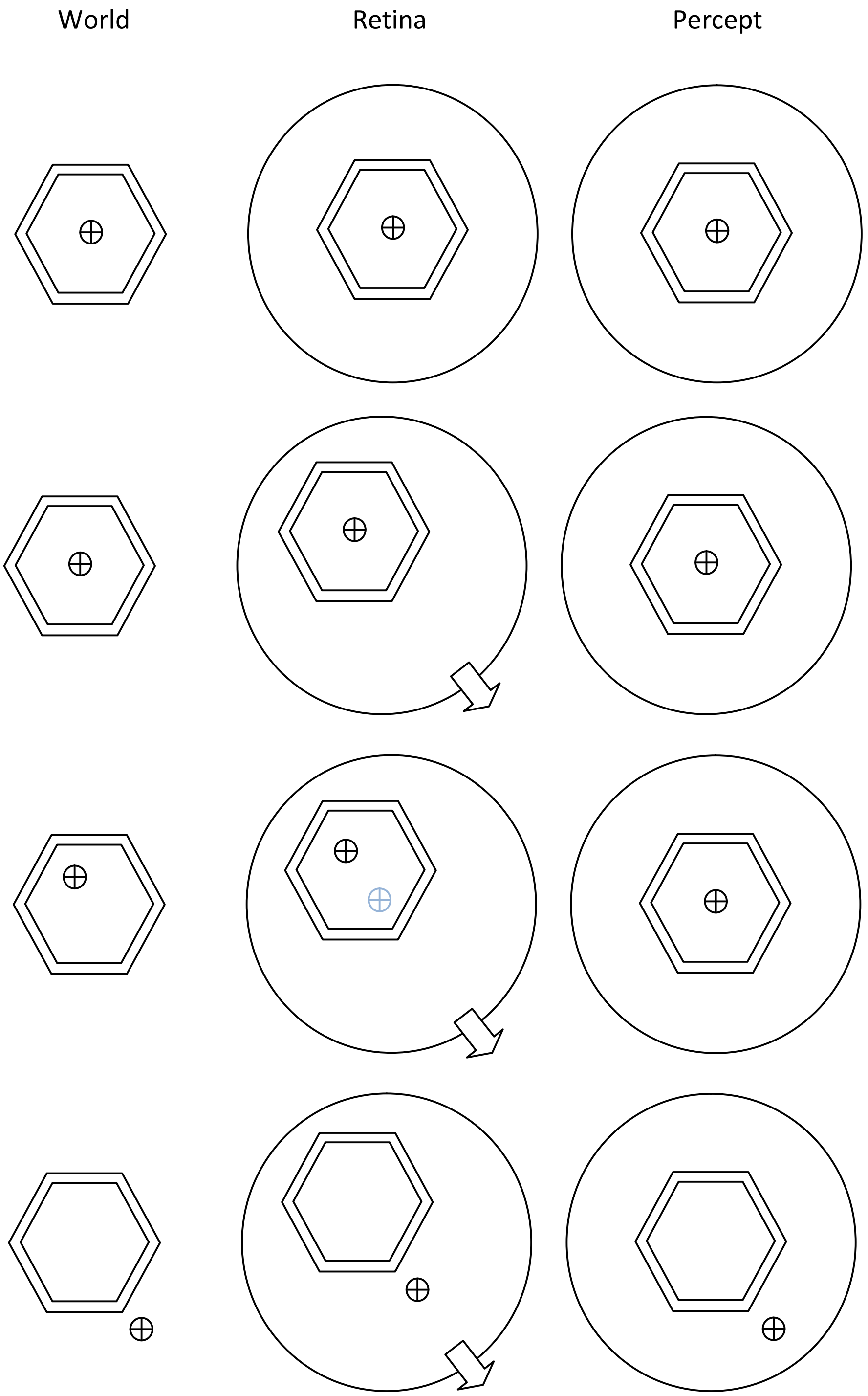

**Figure 9:** First row: reference capture. Second row: eye motion, stimulus gain = 0.0. Third row: eye motion, stimulus gain < 0.0 (also applies to stimulus gain > 0.0 and < 1.0, faint stimulus pattern). Fourth row: eye motion, stimulus gain > 1.0

**Mapping Behavior Implications, Background-Present:**

The mapping behavior induced by each presentation condition illustrated in Figure 9 is as follows:

First row retina is captured to canvas with identity mapping, stimulus to background relationship same as world.

Second row: retina is captured to canvas with single mapping (both background and stimulus use same translation) that restores reference relationship and position of background and stimulus.

Third row retina is captured to canvas with two different mappings: background and stimulus independently restored to reference relationship and position. Both mappings translate same direction but different distances. Stimulus percept is motionless (or nearly motionless, as will be discussed later).

Fourth row: retina is captured to canvas with same mapping for background and stimulus which restores background to reference position and preserves world relationship of stimulus to background. Stimulus motion percept is world motion.

**Functional Behavior of Circuit**

The neural circuitry responsible for visual stabilization must have several capabilities:

1) Mapping background pattern moving on retina to a constant location on canvas
2) Mapping one or more stimulus patterns **moving on the retina not opposite*** to eye motion to their locations relative to the background on the stabilized background pattern on the canvas
3) Mapping one or more stimulus patterns **moving on the retina opposite*** to eye motion but largely independent of motion amplitude relative to background motion on the retina to fixed locations relative to the background pattern on the canvas.

In order to accomplish (2) the same mapping parameters (direction and magnitude) which stabilize background motion on the retina are applied to any stimulus **moving not opposite*** to eye motion.

In order to accomplish (3) mapping parameters different from those which stabilize background motion are applied to any stimulus **moving opposite*** to eye motion such that the stimulus pattern on the canvas remains in a constant relationship to background pattern on the canvas.

Condition (3) clearly requires that a stimulus pattern be separated from the background pattern so that each may be mapped by different direction and magnitude parameters to the canvas. Condition (3) also requires that any stimulus satisfying this condition be removed from the background pattern when it is mapped to the canvas so that it doesn't appear in multiple locations. Both of these operations require one or more isolated reference stimulus patterns in neural arrays that can be used to locate the corresponding pattern in the more complex "scene" on the retina. Condition (3) requires that the stabilization circuitry incorporates mechanisms not present in simple camera-like stabilization, which needs only satisfy condition (2).

The main circuit component is a compact mesh of axons and dendrites, Figure 10a (right), that concurrently generates a range of translation (shift) mappings of the the activation pattern input from the retina and compares each with the activation pattern input from one of the reference arrays. A pair of neurons (shown as double headed arrows) is associated with each mapping. One generates a signal proportional to the degree of match between the particular mapping of the retinal pattern and the reference array pattern. The signals from all of these compete in a simple first-to-arrive winner-take-all circuit [5], and the selected one activates the other neuron of its pair, which gates the optimal mapping of the retinal input to the canvas. Axon-dendrite level behavior of this circuit is discussed later in this paper. The above behavior is depicted in sequential form in Figure 10a (left) and used for simplicity in all the descriptions of the stabilization process below and in Figure 1 earlier.

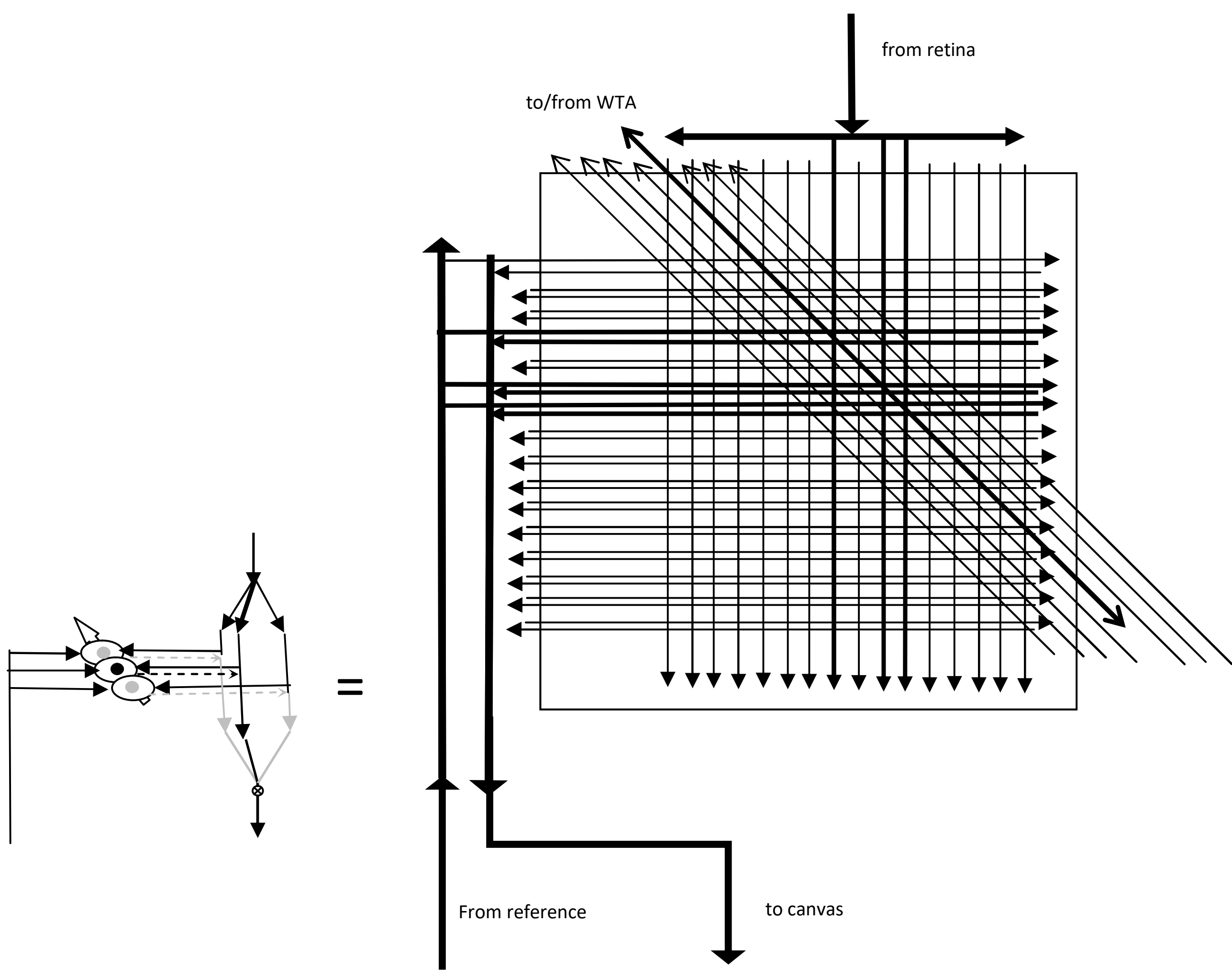

**Figure 10a**

This following section unpacks the actions of the circuit for each phase of the stabilization. The arrays are abbreviated: R for retinal array, BR for background reference array, SR for stimulus reference array. For simplicity all arrays are assumed to have the same dimensions and positions of background and stimulus within them are assumed to be described by the same parameter $\boldsymbol{p}_t$, where vector $\boldsymbol{p}$ implies coordinates, as in $[p_x, p_y]$, and $t$ indicates the time step. The position of background in any array is indicated B$\boldsymbol{p}_t$, and position of stimulus in any array is indicated S$\boldsymbol{p}_t$.

The ellipses with dots each indicate a comparison or match operation between a particular mapping of R and either BR or SR. In other words, if a stimulus is located at $\boldsymbol{p}_1$ in retinal array R at step $t$=1, and the reference stimulus (same pattern) is located at $\boldsymbol{p}_1'$ in stimulus reference array SR at step $t$=1, then the translational mapping $\boldsymbol{p}_1$-> $\boldsymbol{p}_1$" will establish a match, evoking a response from the corresponding compare operator.

**Isolated Background Reference Capture**

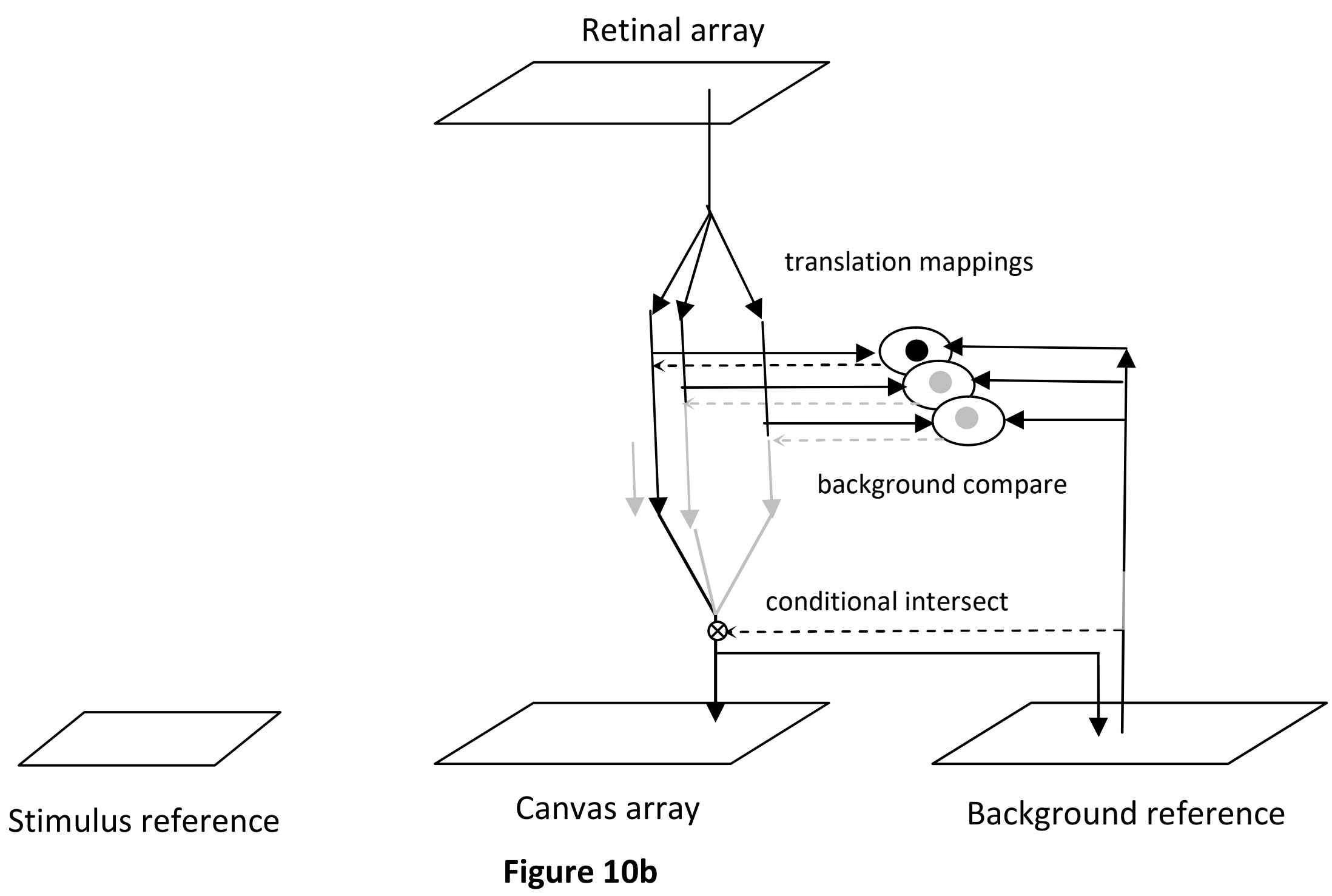

**Figure 10b**

$t_0$: At the start of each fixation the full retinal pattern (stimulus and background) in R is moved to BR and to C via identity mapping. In other words, leaving full retinal image in BR and C in same coordinates as it is in R, background at position B$\boldsymbol{p}_0$, stimulus at position S$\boldsymbol{p}_0$.

$t_1$: Due to eye motion background now at position B$\boldsymbol{p}_1$, stimulus now at position S$\boldsymbol{p}_1$ in R. Assume stimulus has moved relative to background on retina. Therefore B$\boldsymbol{p}_1$ – B$\boldsymbol{p}_0$ (vector operation) is different from S$\boldsymbol{p}_1$ – S$\boldsymbol{p}_0$, i.e. in $t_1$ stimulus has moved by (S$\boldsymbol{p}_1$ – S$\boldsymbol{p}_0$) – (B$\boldsymbol{p}_1$ – B$\boldsymbol{p}_0$) relative to background compared to $t_0$ .

Since background pattern is assumed to remain mostly unchanged, its new position B$\boldsymbol{p_1}$ in R mapped by translation B$\boldsymbol{p_1}$ – B$\boldsymbol{p_0}$, closely matches contents of BR. Mapped R is intersected with contents of BR. The intersection operation erases those elements in mapped R which do not match elements in BR, thereby erasing the stimulus pattern. The resulting isolated background pattern replaces contents of BR, still at position B$\boldsymbol{p_0}$. B$\boldsymbol{p_0}$ will be the stabilized position of the perceived background for the duration of the fixation.

**Stimulus Reference Capture**

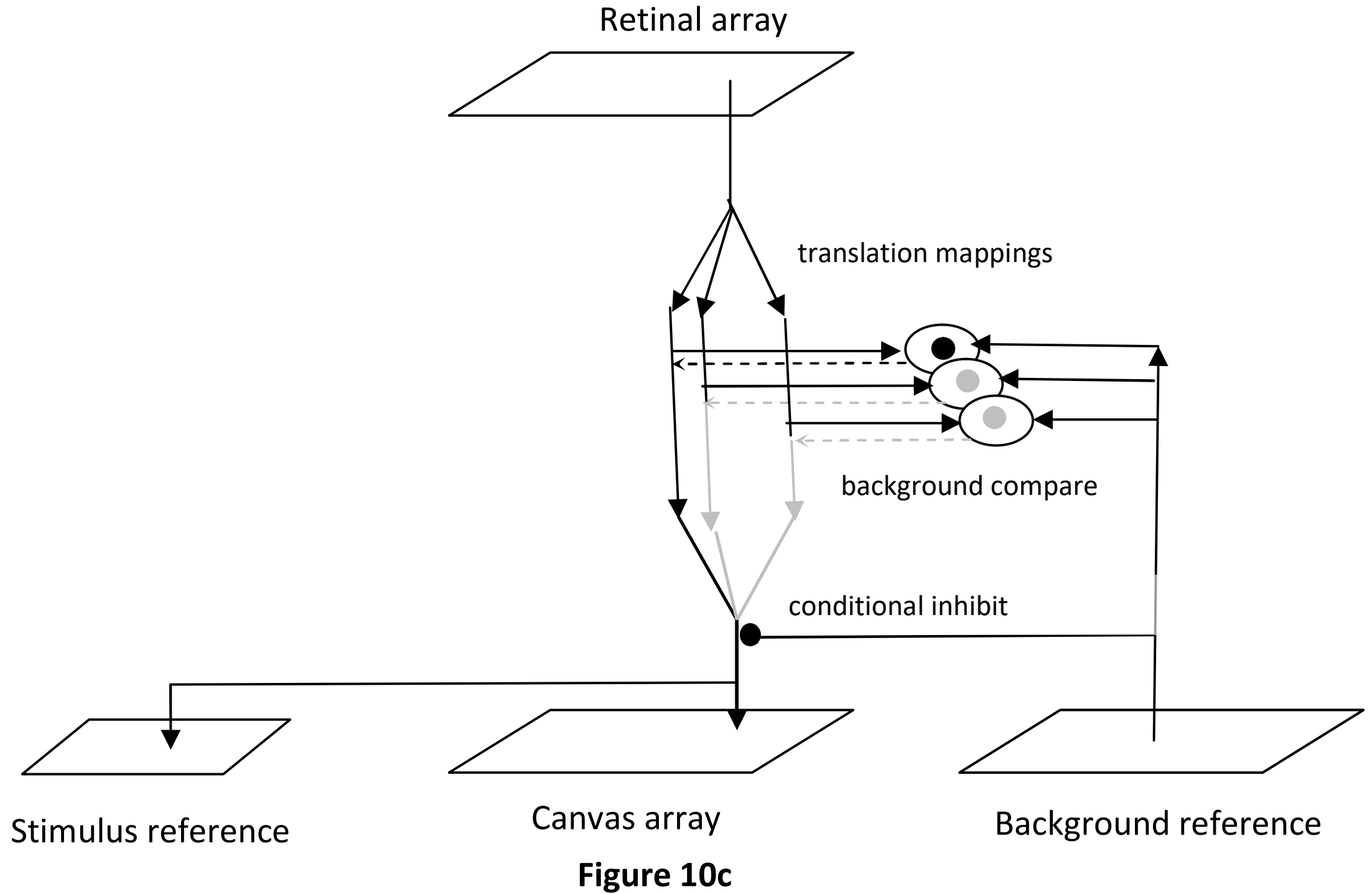

**Figure 10c**

$t_2$: The background pattern has now shifted in R due to eye motion to B$\boldsymbol{p_2}$ and the stimulus has shifted to S$\boldsymbol{p_2}$'. The stimulus and background patterns in R must be shifted by mapping B$\boldsymbol{p_2}$ – B$\boldsymbol{p_0}$ , since this is the mapping that matches the background in R with BR. This is followed by subtraction by inhibition of background pattern in BR. This operation leaves isolated stimulus pattern at position S$\boldsymbol{p_2}$'– (B$\boldsymbol{p_2}$- B$\boldsymbol{p_0}$), which is moved to SR, i.e. shifted by the amount of its motion on R relative to the background pattern. S$\boldsymbol{p_2}$'– (B$\boldsymbol{p_2}$- B$\boldsymbol{p_0}$) will be the perceived position of the stimulus during the fixation when its motion on the retina is opposite eye motion.

Now the isolated background and stimulus reference patterns have been captured in the associated arrays.

**Canvas When Stimulus Motion Induces Percept of Stimulus Stabilization**

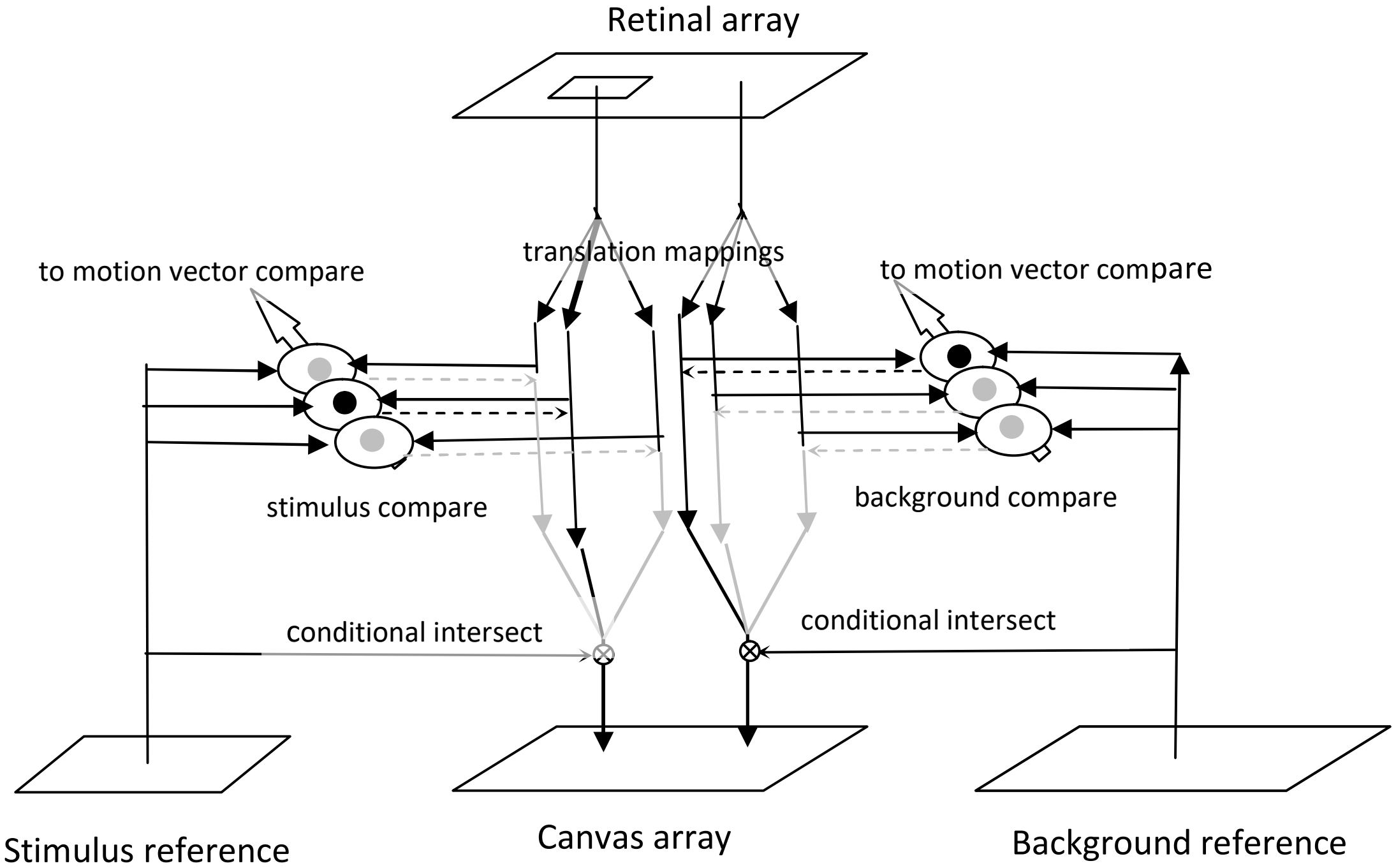

**Figure 10d**

$t_n$: The momentary motion vector of both stimulus is indicated by the change in their associated mappings from $t_{n-1}$ to $t_n$. Each mapping represents both magnitude and direction. Background mapping deltas indicate eye motion, stimulus mapping deltas indicate stimulus motion. Since the decision whether to stabilize is independent of magnitude, all mapping deltas having the same direction (opposite eye motion) constitute the condition to stabilize the perceived stimulus position as stabilized. This condition is determined by the motion vector compare circuitry.

If the stimulus stabilization criterion is met, isolated patterns in SR and BR are transferred (either concurrently or sequentially*) into C, generating the percept of background at $\boldsymbol{p}_0$ and stimulus at $S\boldsymbol{p}_2'$– ($B\boldsymbol{p}_2$- $B\boldsymbol{p}_0$). Notice that all the *t*s (subscripts) are constant for any *n*: this remains the percept so long as the stimulus stabilization condition is met during the fixation.

As with the experimental results, any position of background on R and any position of stimulus on R will result in the same constant positions of both on C. Percept is full stabilization (ignoring latency induced effects)

*Note: B and R patterns may be processed simultaneously in this architecture. Gating of SR to C after BR to C will induce small experimentally observed latencies in S position proportional to stimulus pattern velocity on R.

**Canvas When Stimulus Motion Induces Percept of Veridical Stimulus Motion Relative to Background**

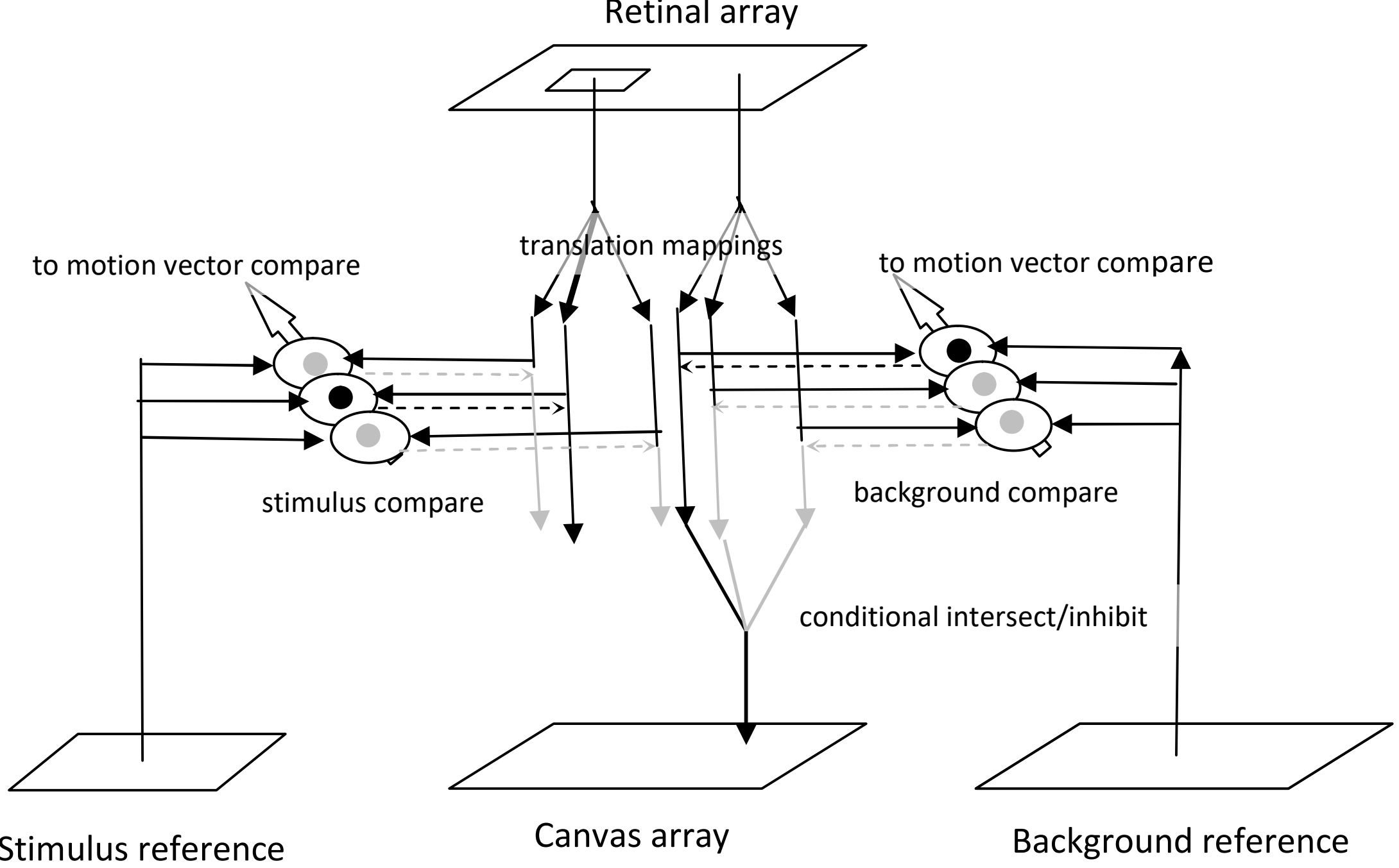

**Figure 10e**

$t_n$: Since the decision NOT to stabilize is independent of magnitude, all stimulus mapping deltas having direction NOT opposite to eye motion constitute the condition to perceived stimulus position relative to background as veridical. This condition is determined by the motion vector compare circuitry.

If this condition is satisfied the stimulus and background patterns in R are mapped by the same mapping, $B\boldsymbol{p}_n – B\boldsymbol{p}_0$ , which stabilizes background in its position in BR at $B\boldsymbol{p}_0$. The percept in C shows the stimulus position as $S\boldsymbol{p}_n'– (B\boldsymbol{p}_n – B\boldsymbol{p}_0)$, which is the same relative positioning as in R. In this condition the perceived stimulus location varies with *n*.

The relationship between sequential stimulus compare neurons signals produce a motion vector indicating both direction and magnitude of motion, not only necessary to determine this mode, but importantly, to report the motion of the stimulus **relative to** the background. Since fixational jitter is approximately random in direction, stimulus motion and eye motion align infrequently, so in natural viewing this is the dominant mode and provides the relevant information necessary for directing attention to targets moving in the world (i.e. relative to the world background). Since all parts of the image on the retina are constantly in motion, the distinction between the moving background and any moving target is not in practice determinable from local retina level velocity detection since the motion across the retina is most likely to be dominated by eye motion rather than target motion. More global “common fate” processing, such as described here, is required.

### Deciding the Stabilization Mode

Downstream from the determination of mappings lies the decision between stimulus stabilization and veridical presentation. This, as experiment revealed, is a function of the diection relationship between momentary eye motion and stimulus motion. At first glance it might seem simplest to use the shift-determining circuitry to directly measure the velocity of stimulus and background motion between the times *t* and *t*-1 by applying it to the current retina pattern and the previously captured one. However, this would imply computing a running vector sum of all the motion increments from the time of reference capture to the current moment. This sum would be needed to drive a mapping circuit to shift the current stimulus and background positions in R all the way back to the initial positions. To keep the percepts stable that running vector sum would need to be quite accurate.

We have to assume that such precise numeric computation is not within the capability of neuronal circuit elements, so summming incremental deltas is not a realizable approach. The circuit described avoids that problem by measuring the current postion relative to the reference position at each *t*, and using that to effect the shift to the reference position. Velocity from *t*-1 to *t* is instead computed as the implied vector difference between sequential mappings from current position to reference position.

But having accepted that vector arithmetic is not readily implemented using neural components, another approach is needed. And the method must involve few neural layers so that the latency between the stimulus mapping decision and the background mapping is minimized. Here the experimentally observed insensitivity of the stabilization decision to motion amplitude provides a clue.

To implement the decision process neuronally, there needs to be a limited set (e.g. 72) of neurons each representing a small range of direction-of-motion from sequential *t*s, ignoring amplitude. The dendrite of such direction-of-motion neurons takes sets of paired inputs from any two match-computing neurons in the mapping circuit, one at time *t*-1 and another at time *t*, whose implicit vector difference, normalized, gives the direction of motion, independent of amplitude. Thus, one direction-of-motion neuron represents all or a large subset of all the combinations of match-computing neurons from sequential *t*s that indicate the same direction of motion, regardless of amplitude.

$$s_i, b_i = \vee^{all}\left(m_{\mathbf{v}}^{t} \wedge m_{\mathbf{v}'}^{t-1}\right) \; s.t. \; \frac{\mathbf{v}_t - \mathbf{v}'_{t-1}}{\left\|\mathbf{v}_t - \mathbf{v}'_{t-1}\right\|} \approx \mathbf{d}_i$$

Where $s_i$ or $b_i$ is the state of the direction-of-motion neuron for stimulus or background direction category *i*, $m_{\mathbf{v}}^{t}$ is the state of the mapping neuron for displacement **v** from reference at time *t*, and $\mathbf{d}_i$ is the center unit vector of direction for category i, in other words an angle category. (This is a specification for the connectivity, not a computation performed by the circuit.)

Only one neuron's latency is involved. No computation is involved. The necessary connectivity can be established either by training, or genetic coding. The dendritic function of such neurons to compute the OR of ANDs is the same as those described later.

The final step in the decision-making requires a layer of neurons which take inputs from the direction-of-motion neurons of both background and stimulus circuitry. These neurons are fired by paired direction-of-motion neurons from background and stimulus which satisfy the stabilization conditions described earlier enable the mappings to the stimulus reference position.

$$S = \left(b_0 \wedge s_{180}\right) \vee \left(b_5 \wedge s_{185}\right) \vee \ldots$$

Where $b_i$ is background direction-of-motion *i* degrees, $s_j$ is stimulus direction-of-motion *j* degrees, *i* and *j* representing ranges around value, $\boldsymbol{S}$ is stabilization state, true or false. Since only one $b_i$ and one $s_j$ are active from the mapping circuit at a given time, this expression is readily implemented by the dendritic computation described below.

Absence of output from this set of neurons enables the direct mapping of stimulus and background in veridical relationship.

Only two chained neurons are involved in the mapping mode decision, keeping the background to stimulus mapping latency to a minimum.

**Stimulus motion away from the axis of eye motion**

The introduction mention that the experiments reported in [1] investigated the percepts induced by stimuli moving at a range of angles to eye motion. Summarizing briefly, it was found that as the stimulus was moved at increasing angles away from the direction exactly opposite to eye motion, the apparent stabilization gradually decreased until the percept was like the percept of the stimulus moving in the same direction as eye motion. Generally, evidence of stabilization was gone by 90 degrees off the maximum stabilization at the axis, thus forming a cardiod-like graph of stabilization versus angle.

There are at least two ways in which the circuitry described above could implement this behavior.

In one possible implementation, when the active direction-of-motion neurons for stimulus and background indicate an off-axis angle relationship, the joint signal would activate a stimulus mapping which would compensate the displacement amplitude relative to background inversely proportional to the angle away from the eye motion axis up to the angle limit observed in the experiments reported in [1].

$$\begin{aligned}
S_{1.0} &= \left(b_0 \wedge \left(s_{180}\right)\right) \vee \left(b_5 \wedge \left(s_{185}\right)\right) \vee \ldots \\
S_{0.9} &= \left(b_0 \wedge \left(s_{170} \vee s_{190}\right)\right) \vee \left(b_5 \wedge \left(s_{175} \vee s_{195}\right)\right) \vee \ldots \\
S_{0.8} &= \left(b_0 \wedge \left(s_{160} \vee s_{200}\right)\right) \vee \left(b_5 \wedge \left(s_{165} \vee s_{205}\right)\right) \vee \ldots
\end{aligned}$$

Where $b_i$ is background direction-of-motion *i* degrees, $s_j$ is stimulus direction-of-motion *j* degrees, *i* and *j* representing ranges around value, $S_k$ is stabilization state, true or false, for amplitude *k* (i.e. degree of stabilization from full to none).

In another implementation, the inverse relationship of angle and stabilization could be implemented by statistical activation of full-stabilization or no-stabilization mappings in inverse proportion over a unit of time so that the average mapping displacement equals the cardiod-specified displacement.

The latter implementation would obviously have a noisier motion characteristic than the first implementation, but the percept characterization technique used in [1] was not refined enough to be able to distinguish between the two.

**Neuronal Circuit Elements:**

While the psychophysics summarized earlier powerfully implies the functional behavior of the neuronal circuit which implements it, the specific architecture of the neuron elements of that circuit is not so constrained. Neurophysiology has not yet provided hints, so the speculation is only limited by required functionality and a rough guess at the computational capabilities of neurons in an extended circuit. Presented next is one possible arrangement. The viability of the functionality of the proposed ciruit, in the sense of what is computed, is proven: it is the same effective algorithm as the original mapping-based stabilization and stimulus delivery mechanism of the AOSLO instrument that was used to obtain the psychophysical evidence [2]. The algorithm used was the Map-Seeking Circuit (MSC), restricted to solve for translations. Two independent analog “neuronal” simulations [ (one in SPICE and one coded from scratch) of the circuit [5], closely related to that outlined below, exhibit the same behavior as the algorithm (MSC) originally employed by the AOSLO instrument for stimulus delivery [2], though this was later replaced by image correlation by FFT-IFFT when GPUs became available and made FFTs fast enough. However, as the equivalent results of MSC (seeking translations only) and FFT-IFFT illustrate, there is often more the one way to compute to obtain the same result, so three or more ways may be as plausible as two. Presented below is the MSC-related approach, implemented “neuronally.”

We start with a neuronal implementation of the matching operation building block of the full neuronal circuit equivalent of Figure 10a(right). It implements the functional equivalent of a dot product between the retinal array shifted by a certain distance and direction and the reference array. The multiplication of the corresponding elements of the two arrays is performed by adjacent synapses such that a signal is only induced on the dendrite at that location if both adjacent synapses are activated. This approximate AND function may be implemented by signal coincidence and non-linear interactions between adjacent synapses. The details are outside the scope of this discussion.

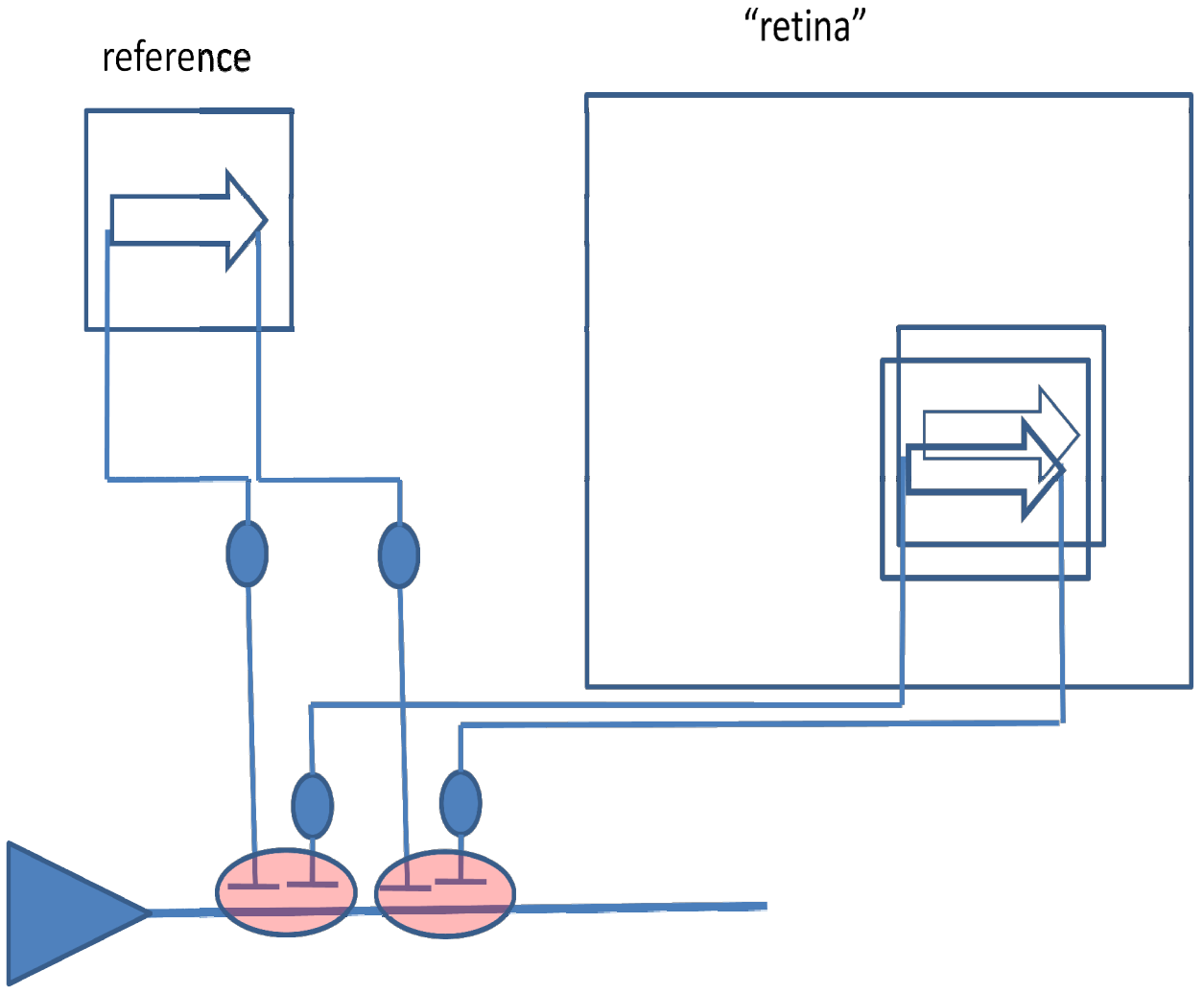

**Figure 11**: Coincident activation of adjacent synapses to produce signal injection to dendrite.

The presence of active input to only one synapse of each of the pair does not result in injection of a signal to the dendrite at that location.

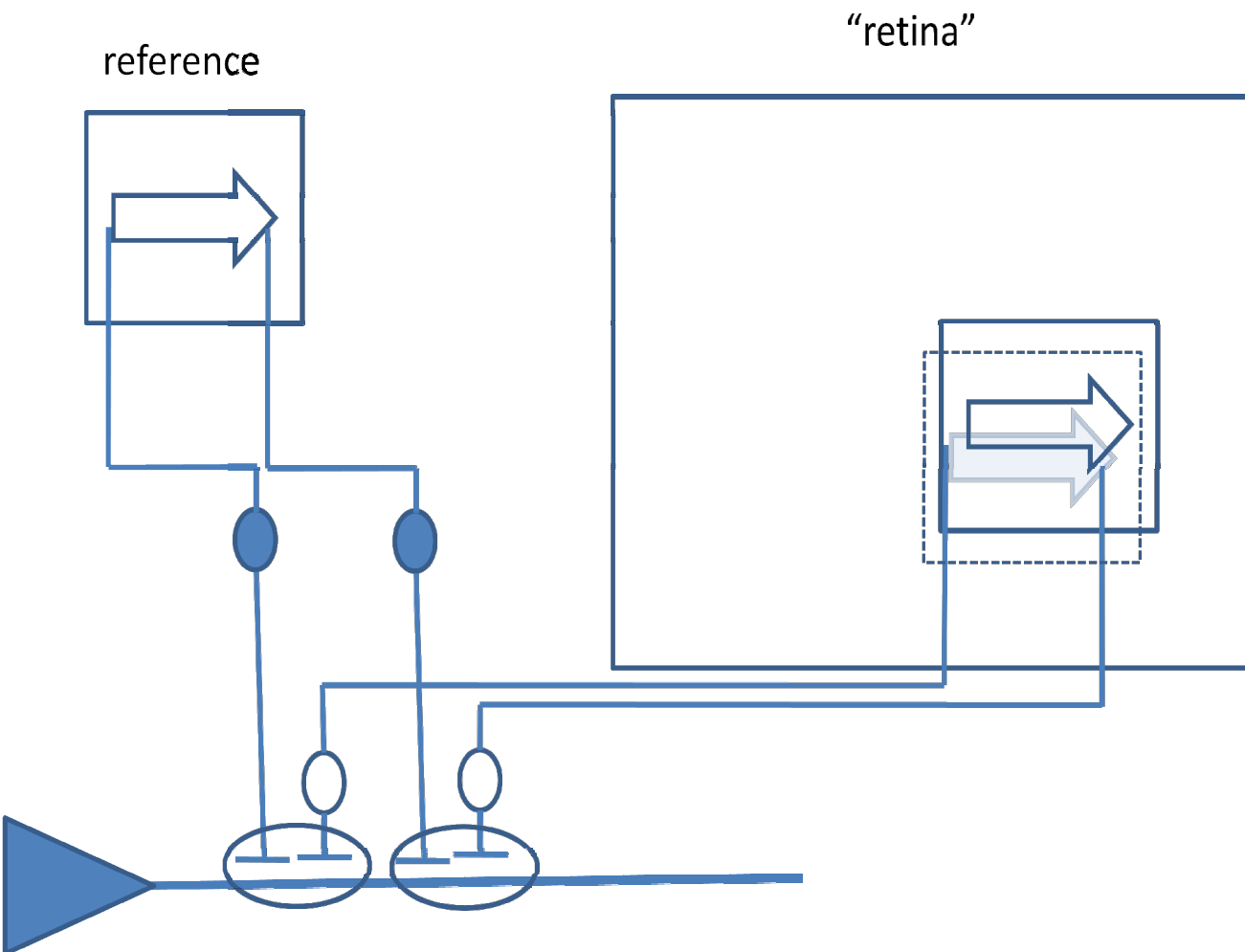

**Figure 12:** Non-coincident activation of adjacent synapses produces no signal injection to dendrite.

Multiple pairs of synapses along the dendrite correspond to the multiple elements of the vector arguments of the dot product, the contribution of those pairs with concurrent signal inputs being approximately summed along the dendrite. The more synapse pairs that contribute to the dendritic signal, the stronger and earlier the leading edge of the pulse induced along the dendrite. The effect is a monotonic but non-linear summation of all the contributing synapse pairs: hence, a rough analog of a dot product [5].

Every translation mapping between the retina array and the reference array has an associated match-computing neuron which “sums” the contribution of the matching element pairs. These match neurons compete via lateral inhibition based on relative pulse timing. The winner is the measure of the best mapping.

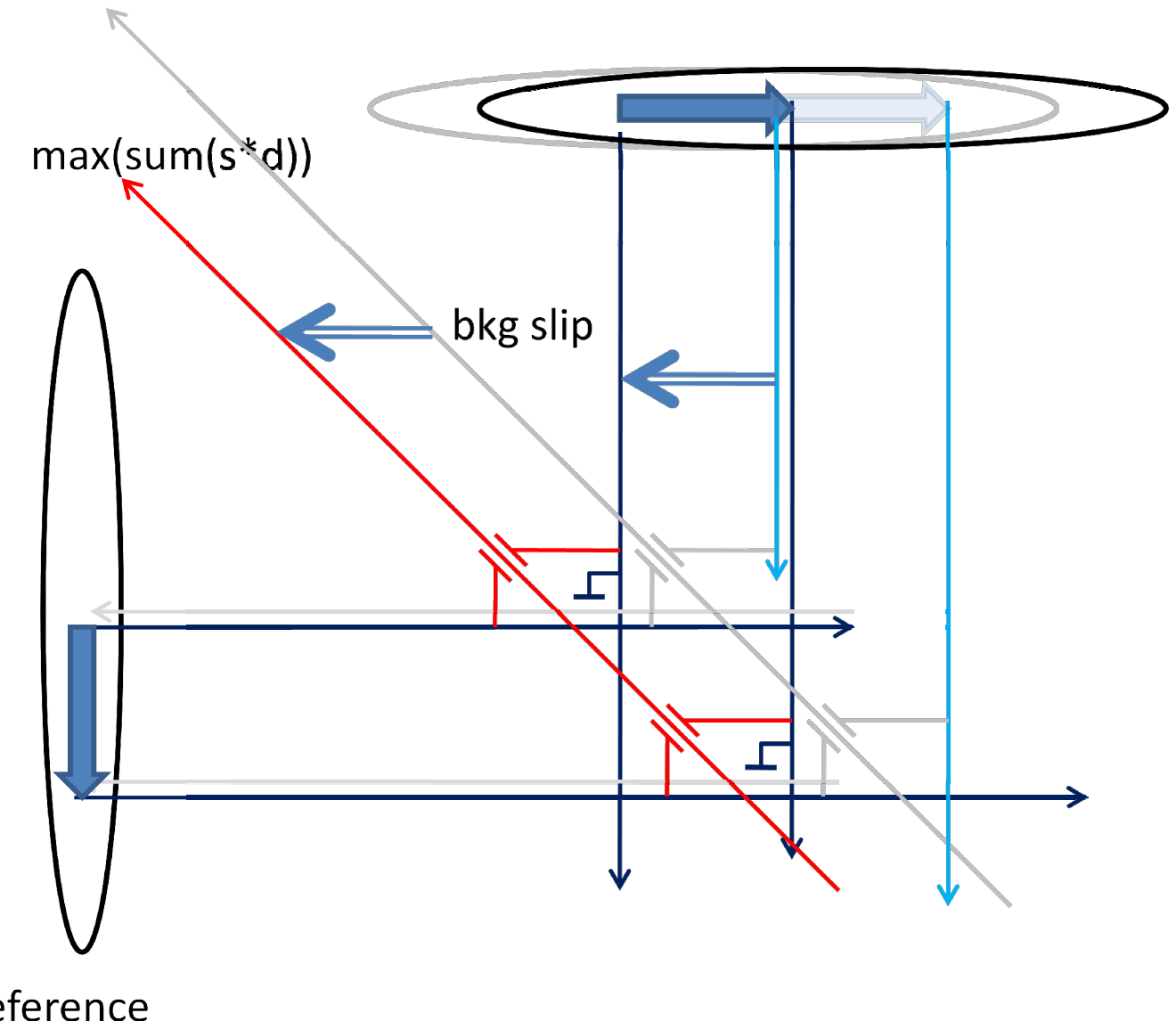

**Figure 13:** Different match neurons compute the degree-of-match for different mappings.

(For a further description and explanation of the competition circuit between these match neurons see [5].)

Once the best mapping has been determined the match neuron activates an associated forward mapping select neuron which transfers the translated pattern of the retina array to the canvas array.

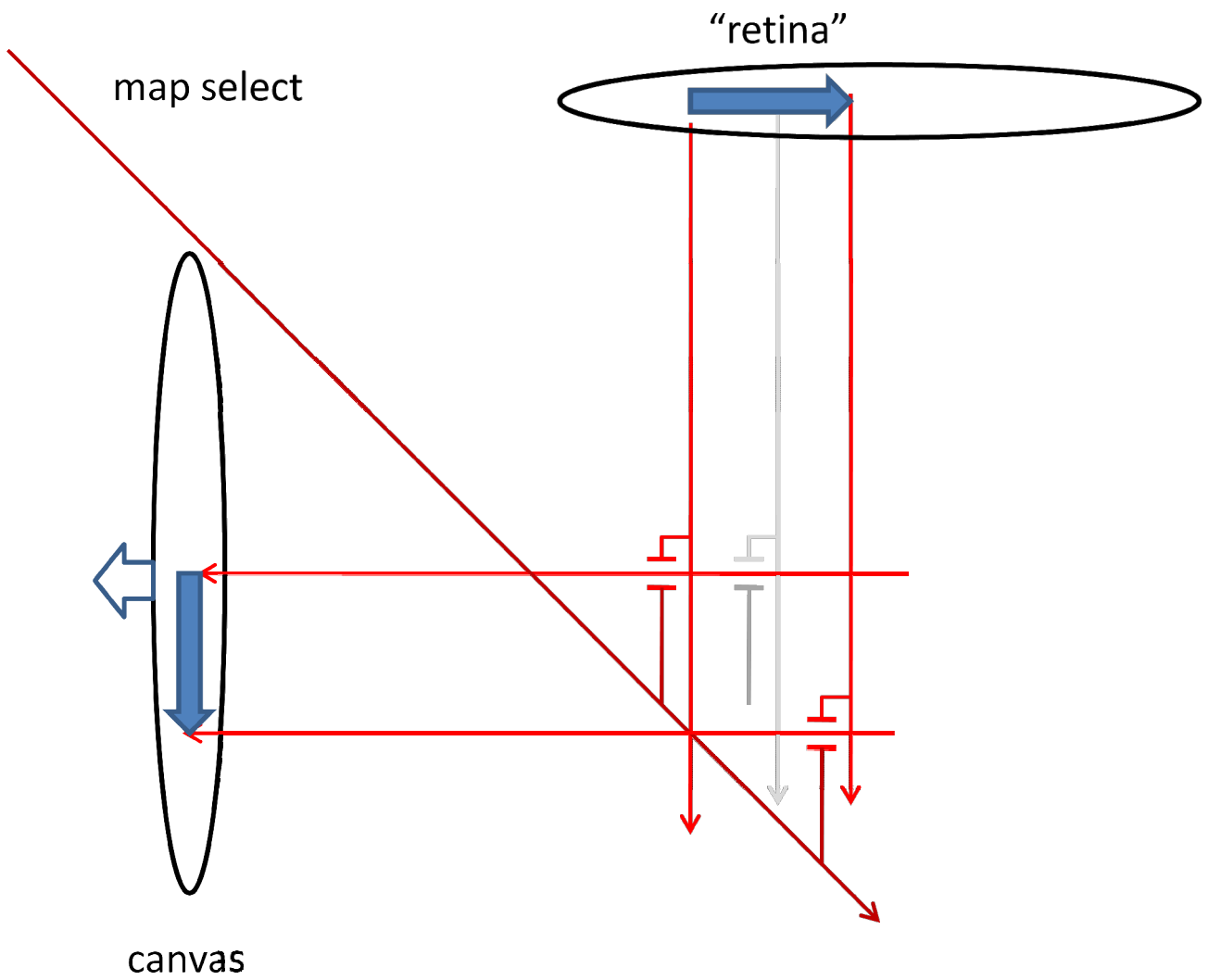

**Figure 14:** A map-select neuron controls the gating of the retinal array to the canvas array.

The entire circuit (for clarity not showing the lateral inhibition circuitry) is shown here. Notice that there is only one pair of neurons for each mapping: the "dot product" computing match neuron and the map select neuron. The pathways from the retina and reference arrays, and to the canvas are shared by all mappings. It is a very efficient use of neuronal resource. (In practice, the number of match neurons per mapping will depend on the number of synapses available on the dendrite and the number of array elements to be matched. The number of map select neurons per mapping will be determined by the number of synapses that can be activated by each axon.)

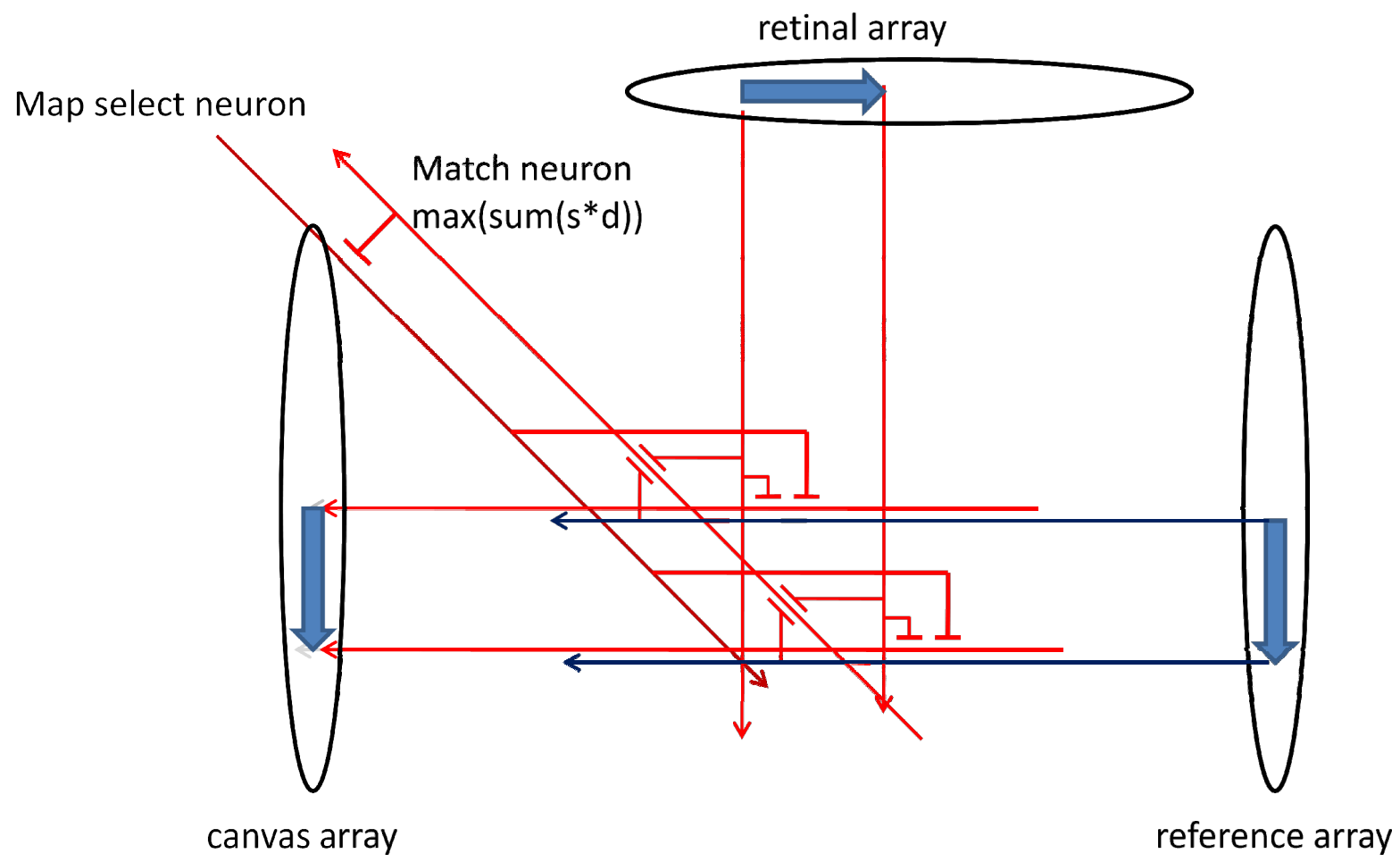

**Figure 15:** The map-select neuron associated with the winning match neuron controls the gating of the retinal array to the canvas array

**Discussion**:

The architecture proposed here is entirely derived from the computations required by the experimentally observed psychophysics. In other words, if one were to write software to simulate the observed behavior, its steps would correspond to the circuit elements presented, though the circuitry parallelizes much of the heavy load. It is also worth noting that all the computing parts of the circuitry are "grown" from combinations of a single component type: a neuron which computes a rough equivalent of a dot product in its dendrite.

There are some questions posed by this hypothesis that the information from the experiments so far performed cannot answer.

1) What is the extent of the mapping area(s)? Since the same mappings are applied to both the background and the stimulus, though independently determined, why are areas of the background near the stimulus **not** transferred to the canvas array along with the stimulus when the stimulus is mapped to its reference position from a different background position? The simplest solution is that the reference array with minor additional circuitry can prune the contribution of the stimulus mapping into the canvas to just the stimulus elements by simple inhibition of the background elements.

2) The mechanics as described above are inherently synchronous. That is, each mapping operation and transfer operation is described as happening in a time step. The dynamics of the biological visual system, however, do not appear to be synchronous, meaning that the retinal array most likely does not update to a "clock" signal, but instead randomly, so that within any small window of time, different portions of the retinal array receive new activation. The dendritic circuitry of the match neurons as described above don't care if only a portion of their synapses have input at any moment so long as the subset of synapses which do have input produce an above-threshold output from the cell. Then the associated mapping select neuron will fire to cause the transfer of those parts of the retinal array that are activated at the same time. This is likely to be a random subset of the full retinal array. This will update only those elements of the canvas array corresponding to the mapped elements of the retinal array that are active at the moment. Therefore even if the cycle rate of a given element of the retinal array is slow, the effective update of the canvas can be fast. (This is analogous to the stabilization process of the AOSLO which remaps one narrow strip of the raster at a time to compute eye motion approximately every millisecond, though the whole raster updates only 60 times per second.)

3) One of the unexpected behaviors reported in the first paper [1] was that all stimuli moving within the stabilization region (i.e. not retricted to moving in the axis of eye motion, but within the "cone" of stabilization) appeared to subjects as motionless with respect to each other even if veridically they were moving with different motions with respect to eye motion. This implies that each stimulus was being stabilized independently and concurrently. This in turn implies that multiple instances of the stabilization mechanism can operate within the viewing field. Since in the experimental configuration reported in [1] one stimulus was in fact the raster and the other a smaller square area within the raster, each moving independently, this might suggest that some stabilization regions can contain others within it, all operating independently. An alternative interpretation is that there are multiple contiguous stabilization regions, and that the containing stimulus is mapped by several regions, all reporting the same relative motion. There is no data from any of the experiments to date to resolve this ambiguity. However, given the temporal requirements of the mechanism it is most likely that stabilization regions are limited in their extent to limit signal propagation time across them, arguing weakly for multiple contiguous regions.

These, and no doubt other, questions cannot be answered without further experiments with test conditions specially designed to elicit distinguishing responses, or better still, until it is possible to monitor the activity of large parts of the circuits involved under controlled test conditions. However, the hypothesis posed above allows purposeful experimental tests to be designed to tease out confirming, revising, or refuting information.

**Notes on origin of proposed circuit:**

The circuit module as presented here is a single stage of a Map-Seeking-Circuit (MSC) [5]. It is bi-directional with pattern arrays providing data at either end. In [5] the configuration provides forward and inverse mappings so that several stages of transformation can be composed. Here, only one class of transformation is required, so only the forward mapping is implemented. As the name implies it seeks the best mapping which registers the patterns in the retinal array to the pattern in the reference array, and then uses that mapping to transfer the shifted retinal array pattern into the canvas array. The mappings here are assumed to be pure translations. In multi-layer configurations the MSC can return compositions of different classes of transformational mappings including rotations and scalings.